\documentclass[a4paper,fleqn]{cas-dc}

\usepackage[numbers]{natbib}
\usepackage{graphicx} 
\usepackage{verbatim}
\usepackage[inline]{enumitem}
\usepackage{comment}
\usepackage[caption=false]{subfig}
\usepackage{multirow}
\usepackage{wrapfig}
\usepackage{amsfonts}
\usepackage{relsize}
\newtheorem{theorem}{Theorem}

\newdefinition{rmk}{Remark}
\newproof{pf}{Proof}
\newtheorem{cor}{Corollary}[theorem]
\newproof{pot}{Proof of Theorem \ref{thm2}}
\newcommand{\subf}[2]{%
  {\small\begin{tabular}[t]{@{}c@{}}
  #1\\#2
  \end{tabular}}%
}
\newcommand{\cmmnt}[1]{}
\def\hyph{-\penalty0\hskip0pt\relax}

\def\tsc#1{\csdef{#1}{\textsc{\lowercase{#1}}\xspace}}
\tsc{WGM}
\tsc{QE}
\tsc{EP}
\tsc{PMS}
\tsc{BEC}
\tsc{DE}

\begin{document}
\newcolumntype{L}[1]{>{\raggedright\arraybackslash}p{#1}}
\newcolumntype{C}[1]{>{\centering\arraybackslash}p{#1}}
\newcolumntype{R}[1]{>{\raggedleft\arraybackslash}p{#1}}

\let\WriteBookmarks\relax
\def\floatpagepagefraction{1}
\def\textpagefraction{.001}
\shorttitle{Evolving Multi-label Classification Rules by Exploiting High-order Label Correlation}
\shortauthors{Shabnam Nazmi, Xuyang Yan, Abdollah Homaifar, Emily Doucette}

\title [mode = title]{Evolving Multi-label Classification Rules by Exploiting High-order Label Correlation}                      

\author[1]{Shabnam Nazmi}[orcid=0000-0001-7511-2910]
\ead{snazmi@aggies.ncat.edu}
\credit{Conceptualization of this study, Methodology, Software}

\author[1]{Xuyang Yan}[orcid=0000-0002-8836-1534]
\ead{xyan@aggies.ncat.edu}

\author[1]{Abdollah Homaifar}[orcid=0000-0003-1179-3221]
\cormark[1]
\ead{homaifar@ncat.edu}

\author[2]{Emily Doucette}
\ead{emily.doucette@us.af.mil}

\address[1]{1601 E Market St., Greensboro, NC, USA, North Carolina A$\&$T State University}
\address[2]{101 West Eglin Blvd. Eglin AFB, FL, USA, The Air Force Research Laboratory - Munitions Directorate}


\cortext[cor1]{Corresponding author}


\begin{abstract}

In multi-label classification tasks, each problem instance is associated with multiple classes simultaneously. In such settings, the correlation between labels contains valuable information that can be used to obtain more accurate classification models. The correlation between labels can be exploited at different levels such as capturing the pair-wise correlation or exploiting the higher-order correlations. Even though the high-order approach is more capable of modeling the correlation, it is computationally more demanding and has scalability issues. This paper aims at exploiting the high-order label correlation within subsets of labels using a supervised learning classifier system (UCS). For this purpose, the label powerset (LP) strategy is employed and a prediction aggregation within the set of the relevant labels to an unseen instance is utilized to increase the prediction capability of the LP method in the presence of unseen labelsets. Exact match ratio and Hamming loss measures are considered to evaluate the rule performance and the expected fitness value of a classifier is investigated for both metrics. Also, a computational complexity analysis is provided for the proposed algorithm. The experimental results of the proposed method are compared with other well-known LP-based methods on multiple benchmark datasets and confirm the competitive performance of this method.

\end{abstract}


\begin{highlights}
\item A multi-label classification model is proposed by extending the supervised learning classifiers through the LP technique
\item The high-order label correlation is exploited to improve the predictive performance and incompleteness challenge is addressed
\item Approximate bounds are derived for the average classifier fitness in terms of the dataset properties 
\end{highlights}

\begin{keywords}
Multi-label classification \sep Label correlation \sep Label powerset \sep Learning classifier systems 
\end{keywords}

\maketitle

\section{Introduction}
In multi-label classification (MLC) tasks, each problem instance is associated with multiple classes at the same time. Emotion identification \cite{almeida2018applying}, image annotation \cite{jing2016multi}, text categorization \cite{jiang2017multi}, semantic scene classification, or gene and protein function prediction \cite{schietgat2010predicting} are examples of such problems. For instance, in text categorization, a document can be classified as History and Biography simultaneously. 

Over the past decade, many multi-label classification algorithms have been proposed to solve the multi-label classification problem in various domains. These algorithms can be categorized into two major groups: problem transformation methods and algorithm adaptation methods \cite{zhang2013review, tsoumakas2009mining}. Problem transformation methods transform the multi-label problem into one or multiple single-label classification problems, e.g., label powerset (LP) \cite{tsoumakas2009mining} and binary relevance (BR) methods \cite{boutell2004learning}. Algorithm adaptation methods modify existing multi-class methods for multi-label problems, such as methods based on $k$NN \cite{zhang2007ml, hanifelou2018knn}, decision tree \cite{vens2008decision, clare2001knowledge}, neural networks \cite{wang2016cnn, zhang2009m}, and support vector machines \cite{elisseeff2002kernel}.

In many real-world multi-label classification problems, a correlation exists between different classes. For instance, a document belonging to the class 'Biography' can also be considered to belong to the class 'History'. Incorporating this information into the classification model could help with obtaining a more accurate classifier. The label correlation can be taken into account through three different strategies, namely \textit{first-order}, \textit{second-order}, and \textit{high-order} \cite{zhang2013review}. The first-order strategy converts the multi-label problem into multiple single-label classification problems and ignores the correlation among labels \cite{zhang2007ml, boutell2004learning}. The second-order strategy considers the pair-wise correlation between labels \cite{elisseeff2002kernel, furnkranz2008multilabel, zhang2006multilabel}, and the high-order strategy, looks at the high order correlation through a subset of labels \cite{read2008pruned, read2011classifier, tsoumakas2010random}.

Many algorithms have been proposed that take into account the second-order correlation often by exploiting the pair-wise relationship between labels. One way to model pairwise correlation is to exploit the co-occurrence pattern between label pairs (e.g. CLR \cite{furnkranz2008multilabel} and LLSF \cite{huang2015learning}) which only consider the positive correlation between labels. On the other hand, LPLC \cite{huang2017multi} and the approach proposed in \cite{nan2018local} exploit local positive and negative pairwise correlation between labels to obtain a MLC model. The PRC algorithm \cite{hullermeier2008label} extends pairwise classification to obtain a ranking procedure based on binary preference relations. Methods developed to capture the high-order correlation are more capable in modeling correlation among labels, but are computationally more expensive and suffer from scalability issues \cite{zhang2013review}. 

High-order approaches mine the relationship between all classes or subsets of classes. Classifier chains (CC) \cite{read2011classifier} is a multi-label classification method that models such relationship by using the vector of class labels as additional sample attributes and transforms the multi-label classification problem into a chain of $q$ binary classification problems. Extensions of the CC algorithm such as probabilistic classifier chains (PCC) \cite{cheng2010bayes}, add a probabilistic interpretation to CC. Also, Bayesian CC \cite{zaragoza2011bayesian} describes the dependency structure of the class labels as a tree. \textcolor{black}{LP is one of the methods that allows for exploiting the high-order label correlation by taking into account label subsets.} Random $k$-labelset (RA$k$EL) \cite{tsoumakas2010random} exploits label correlation in a random way by transforming the problem into an ensemble of multi-class classification problems where each component of the ensemble learns a random subset of the labels through a classifier induced by the LP technique. The ensemble of pruned sets (EPS) \cite{read2008pruned} follows the LP strategy but focuses only on the most important correlations in order to reduce complexity.

Although LP is a straightforward approach to transform multi-label problems into multi-class problems \textcolor{black}{and incorporates label correlation into the learning problem}, it is challenged in two significant ways \cite{zhang2013review}: \begin{enumerate*}[label=(\roman*)]
\item Incompleteness where LP is limited to predict label sets appearing in the training data;
\item Inefficiency, when the number of labels is large, there are too many possible LPs to learn and the training instances for some powersets to learn from are very few which creates class imbalance \end{enumerate*}. RA$k$EL tackles these challenges by combining ensemble learning with LP only on randomly chosen $k$-sized labelsets. 

A learning classifier system (LCS) is a genetic-based machine learning system that combines discovery and learning components to train a rule-based model \cite{holland1992adaptation, urbanowicz2009learning}. The evolutionary component finds new rules and the learning component assigns credit to the rules based on an estimate of their contribution. LCSs are applied in a variety of domains such as biology, computer science, medicine, and social sciences. Three of the major structures developed for LCS are: XCS \cite{iqbal2013evolving}, which operates under the reinforcement learning framework; UCS  \cite{bernado2003accuracy} and ExSTraCS \cite{urbanowicz2015exstracs}, which are developed for supervised learning tasks; and N-LCS \cite{dam2007neural}, which leverages neural networks. In \cite{kim2019exploiting}, several convolutional neural network (CNN) structures are exploited to study the performance of the N-LCSs with CNNs. In \cite{orriols2008genetic}, UCS is successfully used to solve multiple real-world pattern recognition problems. Furthermore, strength-based learning classifiers are adapted to handle multi-label data with weighted labels in \cite{nazmi2017multilabel}, and \cite{nazmi2018multi} investigated the UCS algorithm for its potential in solving multi-label classification problems. 

In this paper, the high-order strategy to handle label correlation through the LP technique is considered and the UCS algorithm is adapted to evolve a rule-based multi-label classification model. The prediction of each rule is a subset of labels induced from the training data. The genetic algorithm (GA) creates new rules by combining two of the existing rules through genetic operators. To reduce the computational complexity, the genetic search is limited to the classifier condition. To overcome the incompleteness of the LP technique on unseen samples, unlike methods that consider random subsets of labels, the proposed method generates new LPs by exploiting the information learned within each problem niche collectively. This approach adopts a similar prediction scheme to a $k$NN method with a dynamic $k$ that aggregates predictions from all the relevant (matching) rules to a given instance. Approximate bounds are derived for the expected value of a classifier's fitness using the average Hamming distance and average Hamming weight bounds. Moreover, a computational complexity analysis is performed for the proposed algorithm. The major contributions of this work are as follows:
\begin{itemize}
    \item A new multi-label classification technique is developed that exploits high-order label correlation by adapting the traditional UCS algorithm to predict multi-labels through the LP technique. Inspired by the $k$NN method, a prediction aggregation is proposed to tackle the incompleteness of the LP technique.
    \item For evaluating the fitness of the classification rules, two strategies are considered. The average classifier fitness using each evaluation strategy is derived in terms of the multi-label data properties and discussions provide insight on the derived bounds. These strategies are also studied through experiments on synthetic and real-world datasets.
    \item Experiments on multiple benchmark datasets are conducted to compare the proposed method with other well-known multi-label classification methods and statistical analyses are performed to analyze the results.
\end{itemize}
The rest of the paper is organized as follows: section \ref{sec: MLC problem} notations and important metrics, section \ref{sec: proposed method} the proposed methods and theoretical analysis, and section \ref{sec: results} the experimental setup and results. Finally, concluding remarks and future work are presented.

\section{Multi-label classification problem}\label{sec: MLC problem}
Let $\mathbb{X}$ denote an input space and let $ \mathcal{L} = \{\lambda_1, \ldots, \lambda_m \}$ be a finite set of class labels. Suppose every instance $\mathbf{x} \in \mathbb{X}$, where $\mathbf{x}\in \mathbb{R}^d$, is associated with a subset of labels $L \subset \mathcal{L}$, which is often called the set of \textit{relevant} labels. The complement set of $L$ is called the \textit{irrelevant} set and is shown by $\bar{L}$. Therefore, $D = \{(\mathbf{x}_1, L_1), (\mathbf{x}_2, L_2), \ldots, (\mathbf{x}_n, L_n)\}$ is a finite set of training instances that are assumed to be randomly drawn from an unknown distribution. The objective is to train a multi-label classifier $h: \mathbb{X} \rightarrow 2^{\mathcal{L}}$ that best approximates the training data and generalizes well to the samples in test data. The function $f(\mathbf{x}, \lambda)$ calculates the score value for class $\lambda$. 

To characterize the properties of a multi-label problem that influences the learning performance, various metrics are proposed in the literature. \textit{Label cardinality} (\ref{eq: card}) is the average number of labels per sample, and \textit{label density}  (\ref{eq: dens}) is the cardinality divided by the number of classes \cite{tsoumakas2007multi}. 

\begin{small}
\begin{align}\label{eq: card}
    Card (D) = \dfrac{1}{n}\sum_{i=1}^{n}|L_i|,
\end{align}
\begin{align}\label{eq: dens}
    Dens (D) = \dfrac{1}{n}\sum_{i=1}^{n}\dfrac{|L_i|}{m} = \dfrac{Card(D)}{m}.
\end{align}
\end{small}
Moreover, in \cite{zhang2009feature} and \cite{tsoumakas2010random}, the \textit{distinct labelsets} (DL) is defined as the number of different label combinations in the dataset:
\begin{small}
\begin{align}
    DL(D) = |L \subset 2^{\mathcal{L}}| (\mathbf{x}, L) \in D|.
\end{align}
\end{small}
In \cite{zhang2009feature}, the \textit{proportion of distinct labelsets} (PDL) is defined as the number of distinct labelsets relative to the number of instances:
\begin{align}\label{eq: pdl}
    PDL(D) = \dfrac{DL(D)}{n}.
\end{align}
\section{Proposed methodology}\label{sec: proposed method}
In this section, the structure of the proposed multi-label learning classifier system (abbreviated as MLR for multi-label classification rules) is explained. A theoretical analysis is presented to show the relationship between the expected fitness of a classification rule in MLR and dataset properties using two fitness evaluation strategies. Besides, the computational complexity of the proposed algorithm is discussed.
\subsection{Algorithm structure}
The proposed algorithm consists of two main components; rule structure and multi-label prediction.
\subsubsection{Multi-label rule structure}
The following shows an example of a rule in multi-label setting that matches $\mathbf{x}$,
\begin{align}\label{eq: rule}
R: \{\text{if } \mathbf{x} \in [\mathbf{c}-\mathbf{s}, \mathbf{c}+\mathbf{s}] \text{, then } \textit{prediction}=\mathbf{y}\},
\end{align}
where $\mathbf{c}$ and $\mathbf{s}$ are vectors of center and spread of a hyper-rectangle respectively encoding the classifier condition $\delta$, and $\mathbf{y}$ is a binary vector with a '1' representing relevant labels and a '0' representing irrelevant labels to $\mathbf{x}$, where $\mathbb{y} \in L$. Every time the \textit{covering} mechanism creates a new rule, it assigns the correct labelset of the training instance to the created rule. The set of all matching rules comprises the \textit{match set} ($[M]$). The experience ($exp$) of a rule is the number of times that it has matched $\mathbf{x}$, and its numerosity ($num$) is the number of copies of it in the \textit{population} ($[P]$) created by GA. It is assumed that the maximum number of the rules allowed in $[P]$ is $N$. In the MLR algorithm, no genetic search is applied to the label space and the off-springs take the exact same predictions as their parents. Moreover, fitness ($\mathcal{F}$) specifies  the relative predictive performance of a rule and is used by GA as a measure of contribution. Fitness of a rule at iteration $t$ can be calculated using different multi-label performance metrics: 
\begin{align}\label{eq: classifier accuracy-em}
    \mathcal{F}_{t}^{em} = (\dfrac{em_{t}}{exp_{t}})^{\nu},
\end{align}
\begin{align}\label{eq: classifier accuracy-hl}
    \mathcal{F}_{t}^{hl} = (\dfrac{1 - hl_{t}}{exp_{t}})^{\nu}.
\end{align}
In the above equations, $em_{t}$ and $hl_{t}$ are the exact match (EM) and the hamming loss (HL) measures of a rule's prediction at iteration $t$, respectively. These values can be calculated at each iteration using $\sum_{i=1}^{exp_{t}} M_i$, for a given multi-label metric $M$. Moreover, $\nu$ is a constant set by the user that determines the strength pressure toward accurate rules \cite{orriols2006revisiting}. 


\subsubsection{Multi-label prediction}
In the original UCS, the predicted class is the one that is predicted by the classifier with the highest fitness value. Let $\mathbf{y}$ to be the prediction of rule $R$, then the predicted multi-label can be obtained as follows:
\begin{align}\label{eq: y max}
    \mathbf{y}_{max} : \{\mathbf{y}|\mathcal{F}=\max \mathcal{F}^l; l = 1, \ldots, |[M]|\},
\end{align}
where $\mathcal{F}^l$ is the fitness of rule $R^l$. We call the algorithm that implements this prediction strategy $\text{MLR}_{max}$. Nonetheless, to overcome the incompleteness of the LP-based learning, labelsets predicted by the locally relevant rules (rules that appear in the $[M]$) are aggregated and each class label is assigned a score value. More specifically, the aggregation of predictions is composed of two steps; $\left. 1 \right)$ to assemble all of the predicted classes into a unified prediction and  $\left. 2 \right)$ to construct a combined score for each class. To perform the former, a simple union of the all the predicted labels in $[M]$ is considered. Let $\mathbf{y}^l$ to be the prediction of rule $R^l$, then the predicted combined multi-label $\bar{\mathbf{y}}$ is:
\begin{align}\label{eq: combinedY}
\bar{\mathbf{y}} : \{\bar{y}_i \hspace{5pt} | \hspace{5pt} \exists R^l : \bar{y}_i\in \mathbf{y}^l;\hspace{5pt}             i&=1,\ldots,m;\\ \nonumber
l&=1,\ldots,|[M]|\},
\end{align} 
where $|[M]|$ is the size of the current match set and works similar to a dynamic $k$ in a $k$NN algorithm. In the second step, a score is calculated for each class from the rules in $[M]$ based on the fitness and density of the rules that cover the subspace containing $\mathbf{x}$. The score of the $i^{th}$ class $\lambda_i$ can be calculated as follows,
\begin{align}\label{eq: score}
s_i = \sum_{\substack{l=1,\ldots,|[M]|, \\ y^l_i = 1}} \mathcal{F}^l \times num^l.
\end{align}
A higher score indicates that $\lambda_i$ is advocated by rules with a larger number of copies or higher fitness or both. We call the algorithm that implements this prediction strategy $\text{MLR}_{agg}$. By employing different bi-partitioning methods, a set of relevant labels can be obtained from the normalized scores.

\subsection{Fitness evaluation of a classifier in MLR}\label{sec: fitness evaluation}
Evaluating the fitness of a classifier using criteria (\ref{eq: classifier accuracy-em}) and (\ref{eq: classifier accuracy-hl}) leads to the evolution of classifiers with different properties. In this section, the objective is to find a relation between the expected fitness ($\overline{\mathcal{F}}$) value of a classifier when evaluated with each criterion in terms of dataset properties. Firstly, $\overline{\mathcal{F}}^{em}$ is derived when EM is used. Then, an upper bound is derived on $\overline{\mathcal{F}}^{hl}$ in terms of the number of the classes $m$ and the number of the distinct labelsets $DL$ in section \ref{sec: upper bound F}. Furthermore, a lower bound is derived on $\overline{\mathcal{F}}^{hl}$ in terms of the number of the classes $m$ and the \textit{label density} value in section \ref{sec: lower bound F}.

Consider a random classification rule $R$ with condition $\delta$ and prediction $\mathbf{y}$. 
Assuming an evenly distributed sample space, for a classifier with hyper-cube condition encoding, the probability that $R$ matches $\mathbf{x}$ is proportional to the hyper-cube volume that it covers \cite{debie2019implications}. Assume $V_R$ and $V_D$ to be the volume covered by $R$ and the input space of $D$, respectively. The probability of matching is as follows,
\begin{align}
    P_m(R) = \dfrac{V_R}{V_D} = \dfrac{\Pi_{i=1}^d r_i}{\Pi_{i=1}^d s_i}
\end{align}
where, $r_i$ and $s_i$ are the span of the classifier condition and the input space of $D$, respectively. Assuming a normalized input space, the average volume covered by a classifier is $r_0^d$, where $r_0$ is a user-defined initialization parameter utilized by the covering mechanism. Furthermore, assume that the average generality of the population is $\tau_{\#}$, which is a function of generalization parameter $P_{\#}$ and the probability of mutation by GA. Thus, we reach the following relation for the probability that $R$ is part of the match set,
\begin{align}
    P(R\in [M]) = r_0^{d\tau_{\#}}.
\end{align}
In MLR, the genetic algorithm selects two classifiers through a roulette wheel selection procedure proportional to their fitness.
For a dataset $D$ with $n$ samples, the average number of samples that $R$ matches ($n_m$) is $r_0^{d \tau_{\#}} \times n$ and $L_m\subset L$ is a set containing the labelsets of these samples. Without the loss of generality, assume that $\nu=1$ in equations (\ref{eq: classifier accuracy-em}) and (\ref{eq: classifier accuracy-hl}). Thus, the expected fitness update for a random classifier using the EM criterion is, 
\begin{align}\label{eq: average F em}
    \overline{\mathcal{F}}^{em} = r_0^{d \tau_{\#}} \times \frac{1}{DL},
\end{align}
which is a function of the distinct number of labelsets in $D$. Using the HL criteria, the average fitness of this classifier can be formulated as,
\begin{align}
    \overline{\mathcal{F}}^{hl} = r_0^{d \tau_{\#}} \times \dfrac{\sum_{n_m} (1- hl)}{n_m}.
\end{align}
Substituting $hl$ with its average value based on the average Hamming distance ($ahd$), the following equation is obtained,
\begin{align}\label{eq: average F}
    \overline{\mathcal{F}}^{hl} = r_0^{d\tau_{\#}}\times (1 - \dfrac{ahd}{m}).
\end{align}
In (\ref{eq: average F}), $ahd$ is the average Hamming distance between the labels of the samples that $R$ matches. To find a relation between $\overline{\mathcal{F}}^{hl}$ and the properties of $D$, we first provide a few definitions. 

A \textit{binary code} is a non-empty subset of the $m$-dimensional vector space over the binary field $F_2$ \cite{fu2001minimum}. Assuming that each label combination in $D$ is observed only once, i.e. $DL = n$, the set of labelsets $L$ of a dataset form a binary code with cardinality $DL$. Similarly, $L_m$ forms a \textit{binary code} with cardinality $N_m$. The average Hamming distance for the \textit{binary code} $L$ is defined as
\begin{align}\label{eq: ahd}
    ahd(L) = \dfrac{1}{DL^2}\sum_{L_i\in L} \sum_{L_j \in L} hd(L_i, L_j).
\end{align}
Moreover, \textit{Hamming weight} ($hw$) is the number of non-zero elements in a binary string \cite{zhang2001relation}. The average Hamming weight ($ahw$) of the \textit{binary code} $L$ is defined as
\begin{align}\label{eq: ahw}
    ahw(L) = \frac{1}{DL} \sum_{L_i\in L} hw(L_i).
\end{align}
To the best of our knowledge, there is no closed-form calculation for the equations (\ref{eq: ahd}) and (\ref{eq: ahw}). Therefore, we proceed with employing an upper bound and lower bound approximations for them and obtain bounds for $\overline{\mathcal{F}}^{hl}$.

\subsubsection{An upper bound on the expected classifier fitness}\label{sec: upper bound F}
In the literature, lower and upper bound approximations for the value of the $ahd$ are proposed \cite{zhang2001relation} that consider special cases for the cardinality of a \textit{binary code}. One straight-forward lower bound on $\textit{ahd}(L)$ is as follows \cite{fu2001minimum},
\begin{align}\label{eq: ahd lower bound}
    \dfrac{m+1}{2} - \dfrac{2^{m-1}}{DL} \leq ahd(L).
\end{align}
This inequality is meaningful only when $DL \geq 2^m/(m+1)$ \cite{zhang2001relation}. Inequality (\ref{eq: ahd lower bound}) suggests that a larger number of distinct label sets in a multi-label dataset increases the upper bound on the average Hamming distance on $L$. For a classifier that matches a subset of the instances, $L_m \subset L$ holds. This means that $N_m \leq DL$ and the $ahd$ for this classifier follows the inequality,
\begin{align}\label{eq: ahd L and Lm}
    ahd(L_m) \leq ahd(L).
\end{align}
In (\ref{eq: ahd L and Lm}), the equality holds when $L_m = L$, i.e., $R$ matches all samples in $D$ which is the case when $R$ is a classifier with an over-general condition. In other words, the prediction made by a classifier with an over-general condition has the maximum average Hamming distance to the labelset of a training instance. Putting (\ref{eq: average F}) and (\ref{eq: ahd lower bound}) together, the following upper bound exists for the expected fitness of such a classifier based on the number of distinct labelsets $DL$ and the number of classes $m$,
\begin{align}\label{eq: Fbar upper bound}
    \overline{\mathcal{F}}^{hl}\leq  r_0^{d\tau_{\#}}(1 - \dfrac{m+1}{2m} + \dfrac{2^{m-1}}{m\times DL}).
\end{align}
According to inequality (\ref{eq: Fbar upper bound}), a larger number of distinct labels imposes a smaller upper bound for the expected value of the classifier fitness.

Based on the definition, $\overline{\mathcal{F}}^{hl} = ave(\mathcal{F})|_{0 \leq hd \leq m}$, while $\overline{\mathcal{F}}^{em} = ave(\mathcal{F})|_{hd=0}$. This means that the classifier fitness $\mathcal{F}^{hl}$ at a given time is considered to be non-zero even when its prediction is not an exact match of the true labelset, i.e., its $hd > 0$. As a result of this more frequent positive fitness evaluation, for a classifier $\overline{\mathcal{F}}^{em} \leq \overline{\mathcal{F}}^{hl}$ holds. This means that in the MLR algorithm when classifiers are evaluated with respect to the HL of their predicted labelset, they are expected to receive a larger fitness update on average. This relation implies that the classifiers that are not very accurate but can partially predict the correct multi-label, are respected as contributing classifiers when the evaluation criterion is HL. With a higher fitness value, these classifiers will have a better chance of receiving reproductive opportunity from GA and remain in the population of rules.

\subsubsection{A lower bound on the expected classifier fitness}\label{sec: lower bound F}
An upper bound for $ahd(L)$ is proposed in \cite{zhang2001relation} that relates it to the value of $ahw$ in (\ref{eq: ahw}), as follows
\begin{align}\label{eq: ahw upper bound}
    ahd(L) \leq 2ahw(L) - \frac{2ahw(L)^2}{m}.
\end{align}
Considering that the $ahw$ of a set of binary variables corresponds to the value of label $Card$, as defined in (\ref{eq: card}), within the multi-label learning framework when $DL= n$, the inequality (\ref{eq: ahw upper bound}) offers the following upper bound the on $ahd$ of the labelsets of $D$,
\begin{align}\label{eq: ahd and card}
    ahd(L) \leq 2Card - \frac{2Card^2}{m}.
\end{align}
Putting (\ref{eq: average F}) and (\ref{eq: ahd and card}) together, and replacing label $Dens$ for $\frac{Card}{m}$, we obtain the following lower bound on the expected fitness of a random classifier based the values of the $Dens$ of a dataset,
\begin{align}\label{eq: Fbar lower bound}
    r_0^{n\tau_{\#}} (2Dens^2 -2Dens + 1) \leq \overline{\mathcal{F}}^{hl}.
\end{align}
According to (\ref{eq: Fbar lower bound}), the lower bound of $\overline{\mathcal{F}}^{hl}$ is a quadratic function of $Dens$ whose minimum occurs at $Dens = 1/2$. Furthermore, according to (\ref{eq: ahd and card}), $ahd$ has its maximum value when $Card=m/2$. Therefore, employing the HL criterion allows the individual classifiers in MLR algorithm to expect the smallest average fitness when the average Hamming distance between the classifier prediction and the correct labelset is expected to be at its largest value.  

\subsection{Computational complexity analysis}
In this section, the computational complexity of the proposed multi-label classification algorithm is analyzed in terms of its major components. Today's learning classifier systems, including UCS and ExSTraCS, consist of many interacting components with complex dependencies. In \cite{urbanowicz2015exstracs}, a comprehensive list of these components along with their functional description is provided. The analysis presented here studies the complexity of these components individually without considering the complex interactions among them.

Table \ref{tab: computational complexity} presents the complexity of training the MLR algorithm in terms of its components, as well as its overall complexity for one training iteration. Here, it is assumed that the genetic algorithm employs a tournament selection with size $t$ and a uniform crossover. Given that the algorithm is to be trained for $it$ iterations, \textit{deletion from population} and the \textit{genetic algorithm} repeat numerous times during training. According to Table \ref{tab: computational complexity}, the computational complexity of training the MLR algorithm is of order $O(N.n.d.m)$, which grows linearly in terms of the number of the rules $N$, the number of training instances $n$, the input dimension $d$, and the number of classes $m$.

\begin{table}[]
    \centering
    \begin{tabular}{lc}
        Operation & Big $O$ \\
        \hline
        \multicolumn{2}{c}{One training iteration} \\
        \hline
        Matching & $O(N.d)$\\
        Parameter update (\textit{exp}, $hl$, ...) & $O(|[M]|.m) \leq O(N.m)$\\
        Fitness calculation & $O(|[M]|) \leq O(N)$\\
        \hline
        \multicolumn{2}{c}{Deletion from population} \\
        \hline
        Deletion weight & $O(N)$\\
        Delete from $[P]$ & $O(N)$\\
        \hline
        \multicolumn{2}{c}{Applying genetic algorithm once} \\
        \hline
        GA & $O(N.d.m.t)$\\
        \hline
        \multicolumn{2}{c}{Overall algorithm complexity} \\
        \hline
        One iteration & $O(N.n.d.m)$
    \end{tabular}
    \caption{Computational complexity of the major MLR components and the overall complexity of MLR.}
    \label{tab: computational complexity}
\end{table}

\section{Results and Discussion}\label{sec: results}
In this section, the benchmark datasets and several classification algorithms that are used in the comparisons experiments are described. Then, multi-label evaluation measures are explained, and the strategies employed for parameter instantiation are reported. Finally, results are presented and discussed.
\subsection{Benchmark Datasets}
For comparison, five real-world datasets are used including, Yeast \cite{elisseeff2002kernel}, Emotions \cite{trohidis2008multi}, Flags \cite{goncalves2013genetic}, COMPUTER AUDITION LAB 500 (CAL500) \cite{turnbull2008semantic}, and Genbase \cite{diplaris2005protein}. Table \ref{tab: ml datasets} shows information about the number of instances, features, and classes for each dataset.

\begin{table*}
    \centering
    \footnotesize    
    \begin{tabular}{llccccccc}
        Dataset & Domain & Inst. & Feat. & Label & Card. & Dens. & DL & PDL\\
        \hline
        Yeast & Biology & 2,417 & 103(n) & 14 & 4.237 & 0.303 & 198 & 0.082\\
        Emotions & Music & 593 & 72(n) & 6 & 1.869 & 0.311 & 27 & 0.045\\
        Flags & Images & 194 & 9(c) + 10(n) & 7 & 3.392 & 0.485 & 54 & 0.278\\
        CAL500 & Music & 502 & 68(n) & 174 & 26.044 & 0.150 & 502 & 1\\
        Genbase & Biology & 662 & 1,186(b) & 27 & 1.245 & 0.046 & 32 & 0.048\\
    \end{tabular}
    \caption{ML datasets. In the Feat. column $n$, $c$, and $b$ refer to numeric, categorical, and binary attributes, respectively.}
    \label{tab: ml datasets}
\end{table*}

\subsection{Experimental setup}
The comparison of the proposed algorithm is performed using the implementations of the following algorithms in \textit{MULAN}\footnote{\hyperlink{http://mulan.sourceforge.net/}{http://mulan.sourceforge.net/}} library under the machine learning framework \textit{WEKA} \cite{hall2009weka}. The compared methods are discussed as below: 
\begin{itemize}
    \item Label powerset methods: vanilla LP, RA$k$EL, ensemble of pruned sets (EPS), hierarchy of multi-label classifiers (HOMER) with balanced $k$ means \cite{tsoumakas2008effective}.
    
    \item Binary relevance methods: the ensemble of classifier chaining (ECC).
    
    \item Algorithm adaptation methods: multi-label $k$-nearest neighbors (ML-$k$NN).
\end{itemize}
In the LP algorithm, the implementation of the decision tree algorithm in \textit{WEKA} is employed as the base learner. The proposed method is implemented in Python using the UCS implementation \cite{urbanowicz2014elcs} as the base algorithm. 

In order to improve the performance and speed up the algorithm execution, the \textit{RF-ML} feature selection strategy \cite{spolaor2013relieff} is employed in this work. \textit{RF-ML} is an extension of the well-known \textit{ReliefF} feature selection strategy to multi-label data that takes into account the effect of interacting features in ML problems without a need to transform the ML problem into a multi-class problem. The result of applying feature selection is either a subset or a ranked list of the original features. In the latter case, only the top thirty percent of the features are used for model training \cite{cai2018feature}.

All experiments are carried out on a 2.70 GHz Windows 10 machine with a 16.0 GB RAM.

\subsection{Evaluation metrics}
In this section, evaluation measures used in the experiments are explained \cite{madjarov2012extensive}. 
\subsubsection{Example-based measures}
$\bullet$ \textit{Hamming loss} computes the percentage of labels whose relevance is predicted incorrectly:
\begin{align}
HL(h) = \dfrac{1}{m}|h(\mathbf{x})\Delta L| \nonumber
\end{align}
where $\Delta$ represents the Hamming distance between the two vectors $h(\mathbf{x})$ and $L$.

$\bullet$ \textit{Accuracy} is the relative number of classes predicted correctly to the union of relevant and predicted labels:
\begin{align}
Acc(h) = \dfrac{|h(\mathbf{x})\bigcap L|}{|h(\mathbf{x})\bigcup L|} \nonumber
\end{align}

$\bullet$ \textit{Precision} is the relative number of classes predicted correctly to the set of relevant labels:
\begin{align} 
Pr(h) = \dfrac{|h(\mathbf{x})\bigcap L|}{|L|} \nonumber
\end{align}

$\bullet$ \textit{Recall} is the relative number of classes predicted correctly to the set of all predicted classes:
\begin{align}
Rc(h) = \dfrac{|h(\mathbf{x})\bigcap L|}{|h(x)|} \nonumber
\end{align}

$\bullet$ $F_1$ measure is the harmonic mean of precision and recall of the predicted labels:
\begin{align}
F_1(h) = \hspace{3pt} \dfrac{2\times Pr \times Rc}{Pr + Rc} \nonumber
\end{align}

\subsubsection{Label-based measures}
For an evaluation measure $M$, a \textit{macro}-measure is computed by evaluating the underlying measure once for each label and calculating their mean value. In contrast, a \textit{micro}-measure aggregates the predictions of all labels and evaluates the measure at the end.
\begin{align}
    M_{micro} &= M(\sum_{i=1}^{m} TP_i, \sum_{i=1}^{m} FP_i, \sum_{i=1}^{m} TN_i, \sum_{i=1}^{m} FN_i)\nonumber
\end{align}

\begin{align}
    M_{macro} &= \dfrac{1}{m}\sum_{i=1}^{m}M(TP_i, FP_i, TN_i, FN_i) \nonumber
\end{align}
where \textit{TP}, \textit{FP}, \textit{TN}, and \textit{FN} stand for true positive, false positive, true negative, and false negative respectively in both equations. Based on these definitions, the \textit{micro} and \textit{macro} averages for $F_1$ measure can be calculated as follows.

$\bullet$ $F_{1_{micro}}$ is the harmonic mean between the \textit{micro-precision} and \textit{micro-recall}:
\begin{align} 
    F_{1_{micro}} = \dfrac{2\times Pr_{micro}\times Rc_{micro}}{Pr_{micro} + Rc_{micro}} \nonumber
\end{align}

$\bullet$ $F_{1_{macro}}$ is the harmonic mean between precision and recall where the average is calculated per label and then averaged across all labels. If $p_j$ and $r_j$ are the precision and recall for $\lambda_j$, then:
\begin{align}
    F_{1_{macro}} = \dfrac{1}{m}\sum_{j=1}^{m}\dfrac{2\times p_j\times r_j}{p_j + r_j} \nonumber
\end{align}
\subsubsection{Ranking-based measures}
$\bullet$ \textit{One Error} computes how many times the top-ranked label is not relevant:
\begin{align}
OE(f) = \begin{cases}
1 & \arg \max_{\lambda\in\mathcal{L}} f(\mathbf{x}, \lambda) \notin L \\
0 & otherwise
\end{cases}
\nonumber
\end{align}
$\bullet$ \textit{Rank Loss} computes the average fraction of label pairs that are not correctly ordered:
\begin{align} \nonumber
RL(f) = \dfrac{\#\{(\lambda, \acute{\lambda})|f(\mathbf{x},\lambda)\leq f(\mathbf{x}, \acute{\lambda}), (\lambda, \acute{\lambda})\in L \times \bar{L} \}}{|L|\times|\bar{L}|}
\nonumber
\end{align}

\subsection{Parameter instantiation}
The parameters of the methods used is the comparison are instantiated following the recommendations from the literature. In cases where a parameter is to be determined from a set of values, the value that corresponds to the maximum $F_1$ measure on each dataset is considered in the experiments. All parameters and threshold values are determined through a train-test split on each dataset.

The number of models in RA$k$EL is set to $min(2m, 100)$ for all datasets \cite{tsoumakas2007random}. The size of the labelsets for RA$k$EL is set to $m/2$ as it provides a balance between computational complexity and performance \cite{tsoumakas2007random, read2011classifier}. The number of neighbors in the ML-$k$NN method for each dataset is selected from the set $(6, 20)$ with a step size of 2. The EPS algorithm requires setting multiple parameters: the strategy parameter $s$, denoted as $A_b$ and $B_b$ for strategy $A$ and $B$ respectively, parameter $b$ which is selected from the set $\{1,2,3\}$ for each strategy, parameter $p$ which is selected by decreasing from 5, and finally, the number of models that is set to 10 \cite{read2008pruned}. The number of models in ECC is also set to 10 to be consistent with other ensemble methods. HOMER requires the number of clusters to be determined which is selected from $(2, 6)$ \cite{tsoumakas2008effective}.

In the proposed method, the maximum number of rules allowed in the model, $N$ is selected from $(1000, 6000)$ by 1000 steps, $P_{\#}$ is the probability of replacing an allele in classifier condition with a hash, which is selected from $[0.1, 0.9]$ with a step size of 0.05, and the threshold by which genetic algorithm is applied is selected from $(5, 50)$ with the step size of 10. Once the parameters are determined, the threshold values for all methods on all datasets are selected from $[0.1, 0.9]$ with step 0.05.

\subsection{Results and discussion}
This experimental study aims at addressing the following questions: \begin{enumerate*}[label=(\roman*)]
\item Which strategy is more effective for evaluating individual classification rules is more effective?
\item What is the effect of employing prediction aggregation over the maximum fitness criteria?
\item How effective is the proposed algorithm in exploiting label correlation compared to the other methods?
\end{enumerate*}

\subsubsection{Classifier evaluation strategy analysis}

The fitness of a single classifier can be evaluated using the EM or the HL measures as shown in (\ref{eq: classifier accuracy-em}) and (\ref{eq: classifier accuracy-hl}). In this section, the effect of employing each criterion on the overall model performance is investigated using synthetic and real-world data. For the synthetic data the framework proposed in \cite{Tomas2014mldatagen} is employed using the hyper-cube strategy. Experiment on each dataset is repeated ten times to reduce the variance of the results, and the model performance is reported in terms of the HL of the model on test data in Tables \ref{tab: hl synthetic data} and \ref{tab: em and hl real-world}.

According to Table \ref{tab: hl synthetic data}, the model performs better on synthetic datasets using $\mathcal{F}_{t}^{em}$, while Table \ref{tab: em and hl real-world} shows that $\mathcal{F}_{t}^{em}$ provides smaller test HL values on real-world datasets. According to the discussions in section \ref{sec: fitness evaluation}, during the training $\mathcal{F}_{t}^{em}$ causes the expected update in the fitness of the classifiers to be smaller compared to employing $\mathcal{F}_{t}^{hl}$. This creates an evolutionary pressure towards classifiers with more specific conditions that cover only a few samples or a very small subspace but tend to be more accurate. Such pressure increases the chance of training rules that overfit the training data, especially on real-world problems as observed in Table \ref{tab: em and hl real-world}. On the other hand, $\mathcal{F}_{t}^{hl}$ prevents over-fitting by preserving the classifiers that are not very accurate but predict partially correct labelsets on every iteration. 

According to Table \ref{tab: em and hl real-world}, employing $\mathcal{F}_{t}^{hl}$ leads to models with better performance on four out of five datasets. A one-tailed Wilcoxon signed ranks test with $\alpha=0.05$ and $N=5$ is applied to the results. The test did not have enough evidence for rejecting the null hypothesis, which means that given the current evidence the performance of the MLR algorithm using either evaluation method is not significantly different. However, in the following comparison experiments $\mathcal{F}_{t}^{hl}$, i.e. strategy (\ref{eq: classifier accuracy-hl}), is employed to guarantee the condition for sufficient generalization and avoid over-fitting.

\begin{table*}
    \centering
    \footnotesize    
    \begin{tabular}{lC{1.3cm}C{1.3cm}C{1.3cm}C{1.3cm}C{1.3cm}C{1.3cm}C{1.3cm}C{1.3cm}}
    \hline
    Dataset & 2-class & 3-class & 5-class & 8-class & 10-class & 15-class\\
    \hline
    EM-based & 0.006 & 0.018 & 0.030 & 0.073 & 0.091 & 0.109\\
    HL-based & 0.009 & 0.021 & 0.035 & 0.095 & 0.133 & 0.139\\
    \hline
    \end{tabular}
    \caption{The average test HL $\downarrow$ of the model using EM and HL as evaluation measures on synthetic data. Ten datasets are generated per class size.}
    \label{tab: hl synthetic data}
\end{table*}

\begin{table*}
    \centering
    \footnotesize
    \begin{tabular}{lC{1.3cm}C{1.3cm}C{1.3cm}C{1.3cm}C{1.3cm}C{1.3cm}C{1.3cm}}
    \hline
    Dataset & Yeast & Emotions & Flags & CAL500 &  Genbase\\
    \hline
    EM-based & 0.2610 & 0.2620 & 0.3101 & 0.2104 & 0.0106\\
    HL-based & 0.2180 & 0.2381 & 0.2923 & 0.1963 & 0.0121\\
    \hline
    \end{tabular}
    \caption{The average test HL $\downarrow$ of the model using EM and HL as evaluation measures on real-world data.}
    \label{tab: em and hl real-world}
\end{table*}

\subsubsection{Comparison with other MLC methods}
In this section, the results of training different ML algorithms on the selected datasets are presented in Tables (\ref{tab: HL}-\ref{tab: one error}). The results are obtained by running a 5-fold cross-validation for each dataset. The numbers within the parentheses are the relative rank of algorithms on a dataset with respect to a metric. The highest average rank is shown in bold for each evaluation metric. 

To study the effect of the prediction aggregation strategy (\ref{eq: score}), two sets of results are reported for the proposed method; the results using the equation (\ref{eq: y max}) as $\text{MLR}_{max}$, and the aggregated predictions after applying a bi\hyph partitioning methods as $\text{MLR}_{agg}$. In this study, the reported aggregated performances are better results after applying \textit{One} \textit{Threshold} and \textit{Rank Cut} \cite{ioannou2010obtaining} on the combined predictions using (\ref{eq: combinedY}) and (\ref{eq: score}). Note that, $\text{MLR}_{max}$ scores all classes equally and as a result no ranking is available for classes to be reported in terms of a ranking-based measure.

\begin{figure*}
\centering
\begin{tabular}{cc}
\subf{\includegraphics[width=70mm]{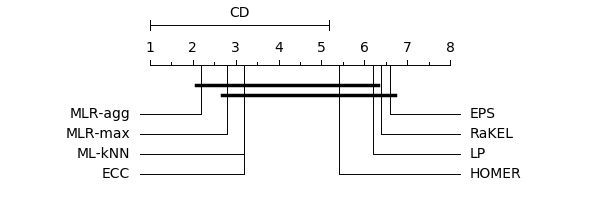}}
     {(a) Hamming Loss}
&
\subf{\includegraphics[width=70mm]{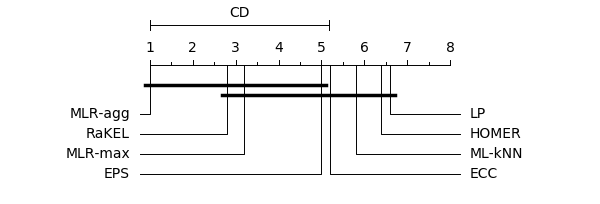}}
     {(b) Accuracy}
\\
\subf{\includegraphics[width=70mm]{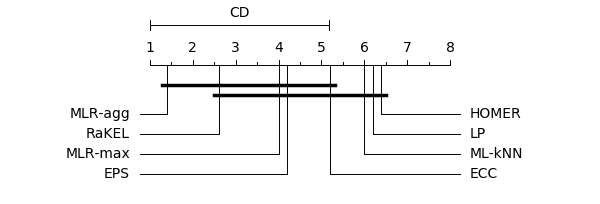}}
     {(c) $F_1$}
&
\subf{\includegraphics[width=70mm]{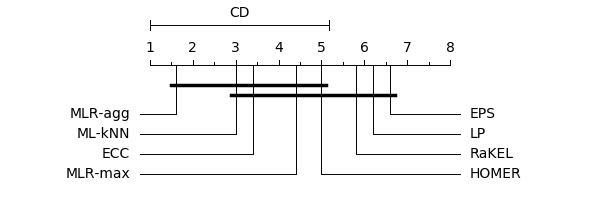}}
     {(d) Precision}
\\
\subf{\includegraphics[width=70mm]{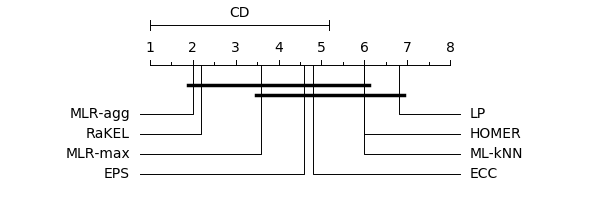}}
     {(e) Micro-$F$}
&
\subf{\includegraphics[width=70mm]{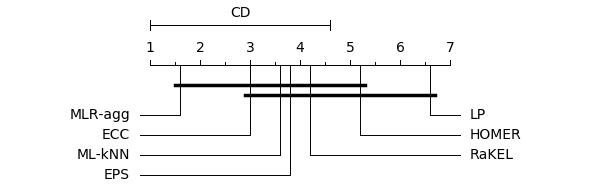}}
     {(f) One Error}
\\
\subf{\includegraphics[width=70mm]{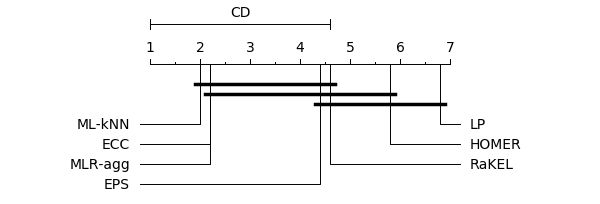}}
     {(g) Rank Loss}
\end{tabular}
\caption{Comparison of MLR with aggregated and max predictions against other algorithms with the Bonferroni-Dunn test with $\alpha=0.05$.}
\label{fig: cd diagrams}
\end{figure*}

To analyze the relative performance of different algorithms, the Friedman test \cite{demvsar2006statistical} is employed. The Friedman test is a non-parametric statistical test to compare multiple algorithms trained on multiple datasets based on their average ranks. According to Table \ref{tab: fridman test results}, the null hypothesis is rejected for all evaluation metrics except for \textit{Recall} and $F_{1_{macro}}$ metrics, suggesting that the performance of the methods are significantly different for all other metrics. Consequently, a post-hoc test \cite{demvsar2006statistical} is applied to investigate the relative performance among algorithms. For this purpose, the Bonferroni-Dunn test \cite{demvsar2006statistical} is employed for $k=8$, i.e. the number of algorithms compared, and $N=5$, i.e., the number of datasets, with a significance level of 0.05. Figure \ref{fig: cd diagrams} shows the critical distance (CD) diagrams for each evaluation metric. \textcolor{black}{The top line in the diagram is the axis along which the average rank of each ML classifier is plotted, from the lowest ranks (best performance) on the left to the highest ranks (worst performance) on the right. In each sub-figure, groups of algorithms that are not statistically different (their average rank is within one CD) from one another are connected.} Following observations are made based on the presented experiments:
\begin{itemize}
    \item According to Tables (\ref{tab: HL}-\ref{tab: rank loss}), MLR algorithm using the prediction aggregation strategy (\ref{eq: combinedY}) has a higher average rank than the maximum prediction strategy (\ref{eq: y max}) in terms of all metrics, which confirms the effectiveness of aggregating the predictions.
    \item $\text{MLR}_{agg}$ has the highest average rank in terms of the six evaluation metrics out of nine and has an outstanding performance in terms of \textit{Accuracy}, \textit{Precision}, and $F_1$ measures. In terms of the \textit{Recall} metric, $\text{MLR}_{agg}$ and RA$k$EL both has the same highest average rank. 
    \item When compared based on the DL value of the benchmark datasets, $\text{MLR}_{agg}$ algorithm has the best performance in terms of six measures on CAL500 dataset which has the highest possible DL value. \textcolor{black}{This result shows that the proposed algorithm is capable of addressing the incompleteness challenge of the LP by predicting unseen labelsets more effectively.}
    \item According to Figure \ref{fig: cd diagrams}, the proposed $\text{MLR}_{agg}$ has significantly better performance than vanilla LP in terms of five metrics. It also offers significant improvement over HOMER on \textit{Accuracy} and $F_1$ score measures, and ML-$k$NN on \textit{Accuracy} measure.
\end{itemize}

\begin{table}
    \centering
    \footnotesize
    \begin{tabular}{lcc}
    \hline
    Evaluation metric & $\mathcal{F}_{F}$ & Critical value ($\alpha=0.05$)\\
    \hline
    Hamming Loss & 19.40 & 14.06\\
    Accuracy & 22.73 & \\
    $F_1$ & 19.00 & \\
    Precision & 17.60 & \\
    Recall & 9.87 & \\
    Micro-$F_1$ & 18.53 & \\
    Macro-$F_1$ & 13.33 & \\
    One Error & 16.29 & \\
    Rank Loss & 23.66 & \\
    \hline
    \end{tabular}
    \caption{Summary of the Friedman rank test for $\mathcal{F}_{F}(k=8, N=5)$.}
    \label{tab: fridman test results}
\end{table}

\begin{table*}
    \centering
    \footnotesize
    \begin{tabular}{llC{1.5cm}C{1.5cm}C{1.5cm}C{1.5cm}C{1.5cm}|C{1.5cm}}
    \hline
    \multicolumn{2}{l}{Datasets} & yeast & emotions & flags & CAL500 & genbase & Ave. rank \\
    \hline 
    \multicolumn{2}{l}{RA$k$EL} & 0.2413(5) & 0.2580(7) & 0.2819(7) & 0.2566(7) & 0.0139(6) & 6.4\\
    \multicolumn{2}{l}{LP($J48$)} & 0.2805(8) & 0.2530(5) & 0.2981(8) & 0.2006(6) & 0.0137(4) & 6.2\\
    \multicolumn{2}{l}{ML-$k$NN} & 0.1976(3) & 0.2218(2) & 0.2458(1) & 0.1404(2) & 0.0158(8) & 3.2\\
    \multicolumn{2}{l}{EPS} & 0.2673(6) & 0.2724(8) & 0.2606(4) & 0.2662(8) & 0.0150(7) & 6.6\\
    \multicolumn{2}{l}{HOMER(LP)} & 0.2791(7) & 0.2533(6) & 0.2804(6) & 0.1966(5) & 0.0135(3) & 5.4\\
    \multicolumn{2}{l}{ECC} & 0.2062(4) & 0.2007(1) & 0.2561(3) & 0.1424(3) & 0.0138(5) & 3.2\\
    \multicolumn{2}{l}{$\text{MLR}_{max}$} & 0.1779(1) & 0.2270(3) & 0.2784(5) & 0.1722(4) & 0.0127(1) & 2.8\\
    \multicolumn{2}{l}{$\text{MLR}_{agg}$} & 0.1816(2) & 0.2357(4) & 0.2518(2) & 0.1402(1) & 0.0131(2) & \textbf{2.2}\\
    
    \hline
    \end{tabular}
    \caption{The performance of the ML algorithms in terms of \textit{Hamming Loss} $\downarrow$.}
    \label{tab: HL}
\end{table*}
\begin{table*}
    \centering
    \footnotesize
    \begin{tabular}{llC{1.5cm}C{1.5cm}C{1.5cm}C{1.5cm}C{1.5cm}|C{1.5cm}}
    \hline
    \multicolumn{2}{l}{Datasets} & yeast & emotions & flags & CAL500 & genbase & Ave. rank \\
    \hline 
    \multicolumn{2}{l}{RA$k$EL} & 0.5008(4) & 0.5280(3) & 0.6114(3) & 0.2505(2) & 0.8319(2) & 2.8\\
    \multicolumn{2}{l}{LP($J48$)} & 0.4113(7) & 0.4764(7) & 0.5533(8) & 0.1982(6) & 0.8252(5) & 6.6\\
    \multicolumn{2}{l}{ML-$k$NN} & 0.4995(6) & 0.5082(6) & 0.6197(2) & 0.1947(7) & 0.6873(8) & 5.8\\
    \multicolumn{2}{l}{EPS} & 0.5044(3) & 0.5234(4) & 0.6077(4) & 0.1903(8) & 0.8183(6) & 5.0\\
    \multicolumn{2}{l}{HOMER(LP)} & 0.4063(8) & 0.4662(8) & 0.5627(7) & 0.2050(5) & 0.8279(4) & 6.4\\
    \multicolumn{2}{l}{ECC} & 0.5001(5) & 0.5190(5) & 0.6058(5) & 0.2118(4) & 0.7632(7) & 5.2\\
    \multicolumn{2}{l}{$\text{MLR}_{max}$} & 0.5763(2) & 0.5312(2) & 0.5863(6) & 0.2176(3) & 0.8318(3) & 3.2\\
    \multicolumn{2}{l}{$\text{MLR}_{agg}$} & 0.5780(1) & 0.5431(1) & 0.6431(1) & 0.3029(1) & 0.8322(1) & \textbf{1.0} \\ 
    \hline
    \end{tabular}
    \caption{The performance of the ML algorithms in terms of \textit{Accuracy} $\uparrow$.}
    \label{tab: accurcay}
\end{table*}
\begin{table*}
    \centering
    \footnotesize
    \begin{tabular}{llC{1.5cm}C{1.5cm}C{1.5cm}C{1.5cm}C{1.5cm}|C{1.5cm}}
    \hline
    \multicolumn{2}{l}{Datasets} & yeast & emotions & flags & CAL500 & genbase & Ave. rank \\
    \hline 
    \multicolumn{2}{l}{RA$k$EL} & 0.6218(4) & 0.6359(3) & 0.7348(2) & 0.3955(2) & 0.8459(2) & 2.6\\
    \multicolumn{2}{l}{LP($J48$)} & 0.5140(8) & 0.5587(8) & 0.6625(8) & 0.3216(6) & 0.8511(1) & 6.2\\
    \multicolumn{2}{l}{ML-$k$NN} & 0.6070(6) & 0.5906(6) & 0.7337(3) & 0.3204(7) & 0.6977(8) & 6.0\\
    \multicolumn{2}{l}{EPS} & 0.6284(3) & 0.6368(2) & 0.7169(4) & 0.3014(8) & 0.8362(4) & 4.2\\
    \multicolumn{2}{l}{HOMER(LP)} & 0.5173(7) & 0.5621(7) & 0.6782(7) & 0.3333(5) & 0.8314(6) & 6.4\\
    \multicolumn{2}{l}{ECC} & 0.6077(5) & 0.5911(5) & 0.7147(5) & 0.3420(4) & 0.7689(7) & 5.2\\
    \multicolumn{2}{l}{$\text{MLR}_{max}$} & 0.6791(2) & 0.6105(4) & 0.6860(6) & 0.3476(3) & 0.8350(5) & 4.0\\
    \multicolumn{2}{l}{$\text{MLR}_{agg}$} & 0.6842(1) & 0.6418(1) & 0.7503(1) & 0.4594(1) & 0.8416(3) & \textbf{1.4}\\    
    \hline
    \end{tabular}
    \caption{The performance of the ML algorithms in terms of $F_1$ score $\uparrow$.}
    \label{tab: f-score}
\end{table*}
\begin{table*}
    \centering
    \footnotesize
    \begin{tabular}{llC{1.5cm}C{1.5cm}C{1.5cm}C{1.5cm}C{1.5cm}|C{1.5cm}}
    \hline
    \multicolumn{2}{l}{Datasets} & yeast & emotions & flags & CAL500 & genbase & Ave. rank \\
    \hline 
    \multicolumn{2}{l}{RA$k$EL} & 0.6030(5) & 0.5763(7) & 0.6497(8) & 0.3089(7) & 0.8387(2) & 5.8\\
    \multicolumn{2}{l}{LP($J48$)} & 0.5384(8) & 0.5886(6) & 0.6613(7) & 0.3293(5) & 0.8330(5) & 6.2\\
    \multicolumn{2}{l}{ML-$k$NN} & 0.7222(1) & 0.6441(3) & 0.7361(2) & 0.5850(1) & 0.7140(8) & 3.0\\
    \multicolumn{2}{l}{EPS} & 0.5616(6) & 0.5617(8) & 0.6929(5) & 0.2897(8) & 0.8290(6) & 6.6\\
    \multicolumn{2}{l}{HOMER(LP)} & 0.5484(7) & 0.6075(5) & 0.6748(6) & 0.3455(4) & 0.8355(3) & 5.0\\
    \multicolumn{2}{l}{ECC} & 0.6888(3) & 0.6461(2) & 0.7081(3) & 0.5609(2) & 0.7711(7) & 3.4\\
    \multicolumn{2}{l}{$\text{MLR}_{max}$} & 0.6765(4) & 0.6411(4) & 0.6938(4) & 0.3171(6) & 0.8343(4) & 4.4\\
    \multicolumn{2}{l}{$\text{MLR}_{agg}$} & 0.7213(2) & 0.8317(1) & 0.8302(1) & 0.4677(3) & 0.8501(1) & \textbf{1.6}\\    
    \hline
    \end{tabular}
    \caption{The performance of the ML algorithms in terms of \textit{Precision} $\uparrow$.}
    \label{tab: precision}
\end{table*}
\begin{table*}
    \centering
    \footnotesize
    \begin{tabular}{llC{1.5cm}C{1.5cm}C{1.5cm}C{1.5cm}C{1.5cm}|C{1.5cm}}
    \hline
    \multicolumn{2}{l}{Datasets} & yeast & emotions & flags & CAL500 & genbase & Ave. rank \\
    \hline 
    \multicolumn{2}{l}{RA$k$EL} & 0.6998(4) & 0.5763(7) & 0.8716(1) & 0.5737(1) & 0.8662(1) & \textbf{2.8}\\
    \multicolumn{2}{l}{LP($J48$)} & 0.5409(8) & 0.5886(6) & 0.6664(8) & 0.3279(5) & 0.8275(6) & 6.6\\
    \multicolumn{2}{l}{ML-$k$NN} & 0.5694(6) & 0.6441(3) & 0.7668(2) & 0.2259(7) & 0.6919(8) & 5.2\\
    \multicolumn{2}{l}{EPS} & 0.7802(1) & 0.5617(8) & 0.7537(3) & 0.2238(8) & 0.8608(2) & 4.4\\
    \multicolumn{2}{l}{HOMER(LP)} & 0.5518(7) & 0.6075(5) & 0.7001(6) & 0.3428(4) & 0.8298(5) & 5.4\\
    \multicolumn{2}{l}{ECC} & 0.5890(5) & 0.6461(2) & 0.7351(4) & 0.2545(6) & 0.7711(7) & 4.8\\
    \multicolumn{2}{l}{$\text{MLR}_{max}$} & 0.7414(2) & 0.6411(4) & 0.6930(7) & 0.4000(3) & 0.8388(4) & 4.0\\
    \multicolumn{2}{l}{$\text{MLR}_{agg}$} & 0.7118(3) & 0.8317(1) & 0.7065(5) & 0.5585(2) & 0.8419(3) & \textbf{2.8}\\    
    \hline
    \end{tabular}
    \caption{The performance of the ML algorithms in terms of \textit{Recall} $\uparrow$.}
    \label{tab: recall}
\end{table*}
\begin{table*}
    \centering
    \footnotesize
    \begin{tabular}{llC{1.5cm}C{1.5cm}C{1.5cm}C{1.5cm}C{1.5cm}|C{1.5cm}}
    \hline
    \multicolumn{2}{l}{Datasets} & yeast & emotions & flags & CAL500 & genbase & Ave. rank \\
    \hline 
    \multicolumn{2}{l}{RA$k$EL} & 0.6350(4) & 0.6587(1) & 0.7513(2) & 0.3984(2) & 0.8550(2) & 2.2\\
    \multicolumn{2}{l}{LP($J48$)} & 0.5387(8) & 0.5924(7) & 0.6904(8) & 0.3253(6) & 0.8511(5) & 6.8\\
    \multicolumn{2}{l}{ML-$k$NN} & 0.6070(6) & 0.6270(6) & 0.7488(3) & 0.3185(7) & 0.8034(8) & 6.0\\
    \multicolumn{2}{l}{EPS} & 0.6367(3) & 0.6516(2) & 0.7417(4) & 0.3012(8) & 0.8435(6) & 4.6\\
    \multicolumn{2}{l}{HOMER(LP)} & 0.5438(7) & 0.5922(8) & 0.7145(6) & 0.3402(5) & 0.8524(4) & 6.0\\
    \multicolumn{2}{l}{ECC} & 0.6312(5) & 0.6512(3) & 0.7391(5) & 0.3430(4) & 0.8434(7) & 4.8\\
    \multicolumn{2}{l}{$\text{MLR}_{max}$} & 0.6871(2) & 0.6333(5) & 0.7060(7) & 0.3535(3) & 0.8565(1) & 3.6\\
    \multicolumn{2}{l}{$\text{MLR}_{agg}$} & 0.6908(1) & 0.6511(4) & 0.7562(1) & 0.4611(1) & 0.8541(3) & \textbf{2.0}\\    
    \hline
    \end{tabular}
    \caption{The performance of the ML algorithms in terms of \textit{Micro-F} $\uparrow$.}
    \label{tab: micro-f}
\end{table*}
\begin{table*}
    \centering
    \footnotesize
    \begin{tabular}{llC{1.5cm}C{1.5cm}C{1.5cm}C{1.5cm}C{1.5cm}|C{1.5cm}}
    \hline
    \multicolumn{2}{l}{Datasets} & yeast & emotions & flags & CAL500 & genbase & Ave. rank \\
    \hline 
    \multicolumn{2}{l}{RA$k$EL} & 0.4276(3) & 0.6499(1) & 0.6922(1) & 0.1915(1) & 0.8478(1) & \textbf{1.4}\\
    \multicolumn{2}{l}{LP($J48$)} & 0.3756(6) & 0.5809(8) & 0.6154(7) & 0.1506(3) & 0.8115(3) & 5.4\\
    \multicolumn{2}{l}{ML-$k$NN} & 0.3609(8) & 0.5889(6) & 0.6221(6) & 0.1052(7) & 0.6577(6) & 6.6\\
    \multicolumn{2}{l}{EPS} & 0.4397(1) & 0.6448(2) & 0.6302(4) & 0.0957(8) & 0.8013(4) & 3.8\\
    \multicolumn{2}{l}{HOMER(LP)} & 0.3841(5) & 0.5822(7) & 0.6375(5) & 0.1702(2) & 0.8146(2) & 4.2\\
    \multicolumn{2}{l}{ECC} & 0.3736(7) & 0.6219(4) & 0.6327(3) & 0.1301(5) & 0.7985(5) & 4.8\\
    \multicolumn{2}{l}{$\text{MLR}_{max}$} & 0.4314(2) & 0.6095(5) & 0.5855(8) & 0.1444(4) & 0.5318(7) & 5.2\\
    \multicolumn{2}{l}{$\text{MLR}_{agg}$} & 0.4031(4) & 0.6230(3) & 0.6444(2) & 0.1195(6) & 0.5240(8) & 4.6\\    
    \hline
    \end{tabular}
    \caption{The performance of the ML algorithms in terms of \textit{Macro-F} $\uparrow$.}
    \label{tab: macro-f}
\end{table*}
\begin{table*}
    \centering
    \footnotesize
    \begin{tabular}{llC{1.5cm}C{1.5cm}C{1.5cm}C{1.5cm}C{1.5cm}|C{1.5cm}}
    \hline
    \multicolumn{2}{l}{Datasets} & yeast & emotions & flags & CAL500 & genbase & Ave. rank \\
    \hline 
    \multicolumn{2}{l}{RA$k$EL} & 0.2954(5) & 0.3222(5) & 0.2468(4) & 0.2211(4) & 0.1662(3) & 4.2\\
    \multicolumn{2}{l}{LP($J48$)} & 0.5337(7) & 0.4234(6) & 0.6233(7) & 0.9832(6) & 0.2328(7) & 6.6\\
    \multicolumn{2}{l}{ML-$k$NN} & 0.2371(2) & 0.3104(3) & 0.2522(5) & 0.1195(2) & 0.1768(6) & 3.6\\
    \multicolumn{2}{l}{EPS} & 0.2474(3) & 0.3171(4) & 0.1854(1) & 0.9868(7) & 0.1708(4) & 3.8\\
    \multicolumn{2}{l}{HOMER(LP)} & 0.5130(6) & 0.4252(7) & 0.3549(6) & 0.8146(5) & 0.1647(2) & 5.2\\
    \multicolumn{2}{l}{ECC} & 0.2478(4) & 0.2918(1) & 0.2059(2) & 0.1474(3) & 0.1753(5) & 3.0\\
    \multicolumn{2}{l}{$\text{MLR}_{max}$} & - & - & - & - &  -  & - \\
    \multicolumn{2}{l}{$\text{MLR}_{agg}$} & 0.1514(1) & 0.3034(2) & 0.2318(3) & 0.1156(1) & 0.1571(1) & \textbf{1.6}\\
    \hline
    \end{tabular}
    \caption{The performance of the ML algorithms in terms of \textit{One-error} $\downarrow$.}
    \label{tab: one error}
\end{table*}
\begin{table*}
    \centering
    \footnotesize
    \begin{tabular}{llC{1.5cm}C{1.5cm}C{1.5cm}C{1.5cm}C{1.5cm}|C{1.5cm}}
    \hline
    \multicolumn{2}{l}{Datasets} & yeast & emotions & flags & CAL500 & genbase & Ave. rank \\
    \hline 
    \multicolumn{2}{l}{RA$k$EL} & 0.2172(5) & 0.1977(5) & 0.2461(4) & 0.2450(4) & 0.0952(5) & 4.6\\
    \multicolumn{2}{l}{LP($J48$)} & 0.4102(7) & 0.3331(7) & 0.5405(7) & 0.6578(6) & 0.1924(7) & 6.8\\
    \multicolumn{2}{l}{ML-$k$NN} & 0.1708(2) & 0.1869(3) & 0.1982(1) & 0.1853(2) & 0.0165(2) & \textbf{2.0}\\
    \multicolumn{2}{l}{EPS} & 0.1998(4) & 0.1890(4) & 0.2265(3) & 0.6802(7) & 0.0556(4) & 4.4\\
    \multicolumn{2}{l}{HOMER(LP)} & 0.3599(6) & 0.3107(6) & 0.3577(6) & 0.4279(5) & 0.1463(6) & 5.8\\
    \multicolumn{2}{l}{ECC} & 0.1798(3) & 0.1650(2) & 0.2018(2) & 0.1987(3) & 0.0135(1) & 2.2\\
    \multicolumn{2}{l}{$\text{MLR}_{max}$} & - & - & - & - & - & - \\
    \multicolumn{2}{l}{$\text{MLR}_{agg}$} & 0.1491(1) & 0.1783(1) & 0.2725(5) & 0.1785(1) & 0.0197(3) & 2.2\\
    \hline
    \end{tabular}
    \caption{The performance of the ML algorithms in terms of \textit{Rank Loss} $\downarrow$.}
    \label{tab: rank loss}
\end{table*}

\section{Conclusion}
In this paper, a multi-label classification algorithm by extending supervised learning classifier systems is proposed to exploit the high-order label correlation in order to obtain a more accurate classification model. The proposed method builds classification rules by extending the LP technique and employs a prediction aggregation that works similar to a $k$NN method with a dynamic $k$. Two strategies for evaluating the performance of the individual classifiers during the training is considered and are investigated by deriving approximate bounds on the expected classifier fitness in terms of the number of classes, the number of distinct labelsets, and the label density in the dataset.  

The complexity analysis reveals that the cost of training MLR is linear in terms of the number of instances, number of features, number of classes, and the number of rules in the population. Experiments on the synthetic and real-world datasets suggest that evaluating classifier performance using the Hamming loss measure is more effective in preventing over-fitting than the exact match measure. This result is due to higher expected fitness for classifiers that are partially correct when evaluated using the HL criteria. The proposed method is compared with multiple well-known multi-label classification methods on multiple datasets and has the highest average rank in terms of the seven out of nine measures. Statistical tests on the results show that the MLR algorithm with aggregated predictions outperforms other methods on most of the datasets and shows competitive performance on others. The lower performance of the model in terms of the macro-averaged $F$ score suggests that the model might present poor prediction performance on datasets with imbalanced classes, where it is necessary to correctly predict the infrequently occurring class labels. 

In the future, the impact of other mechanisms such as the genetic operators and deletion will be incorporated into the analysis presented for the performance of the individual classifiers to obtain a more complete analysis of the MLR algorithm. We will also investigate different techniques to improve the performance of the proposed method on imbalanced class datasets.

\section{Acknowledgement}
This work is supported by Air Force Research Laboratory (AFRL) and Office of Secretary of Defense (OSD) under agreement number FA8750-15-2-0116.
\bibliographystyle{cas-model2-names}

\bibliography{cas-refs}

\vskip20pt
\begin{wrapfigure}{L}{0.2\columnwidth}
\includegraphics[width=0.2\columnwidth]{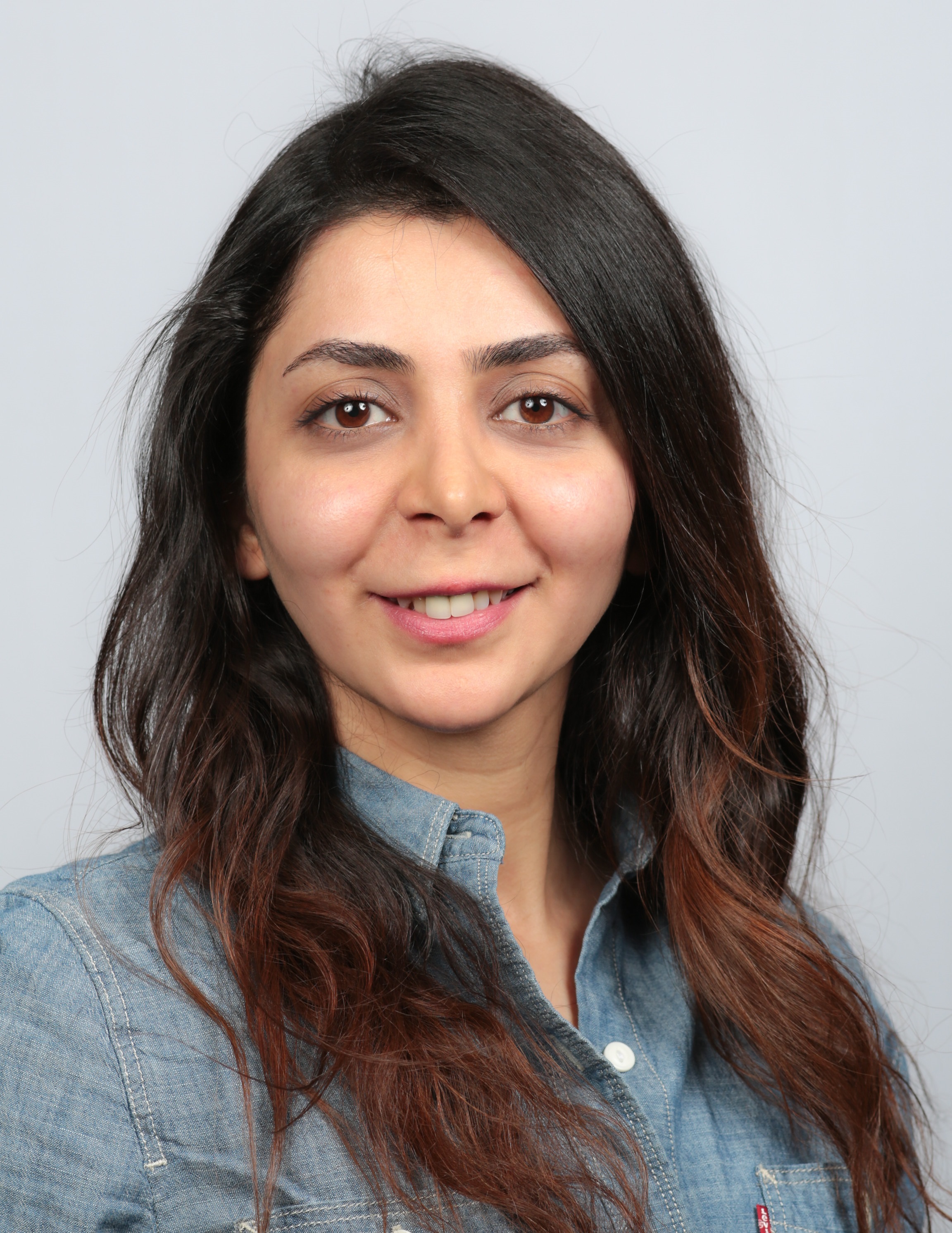}
\end{wrapfigure}
\footnotesize
\textbf{Shabnam Nazmi} received her B.S. degree in Electrical Engineering from K.N.Toosi University of Technology and her M.S. degree in Electrical Engineering from Sharif University of Technology in 2009 and 2012, respectively. She is currently a Ph.D. candidate at the Department of Electrical and Computer Engineering, North Carolina A$\&$T State University. Her research interests include multi-label classification and its application on test and evaluation of autonomous vehicles, genetic-based machine learning, and learning from streaming data.

\begin{wrapfigure}{L}{0.2\columnwidth}
\centering
\includegraphics[width=0.2\columnwidth]{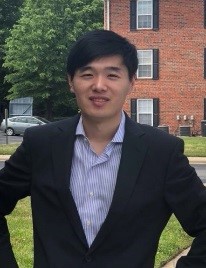}
\end{wrapfigure}
\textbf{Xuyang Yan} received his B.S. degree in Electrical Engineering from North Carolina Agricultural and Technical State University (NC A$\&$T) and Henan Polytechnic University in 2016. In 2018, he earned his M.S. degree in electrical engineering at NC A$\&$T. He is currently pursuing his Ph.D. degree in electrical engineering at NC A$\&$T. His research interests include extracting knowledge from streaming data, analyzing the emergent behaviors of large-scale autonomous systems and the application of machine learning techniques in robotics.

\begin{wrapfigure}{L}{0.2\columnwidth}
\centering
\includegraphics[width=0.2\columnwidth]{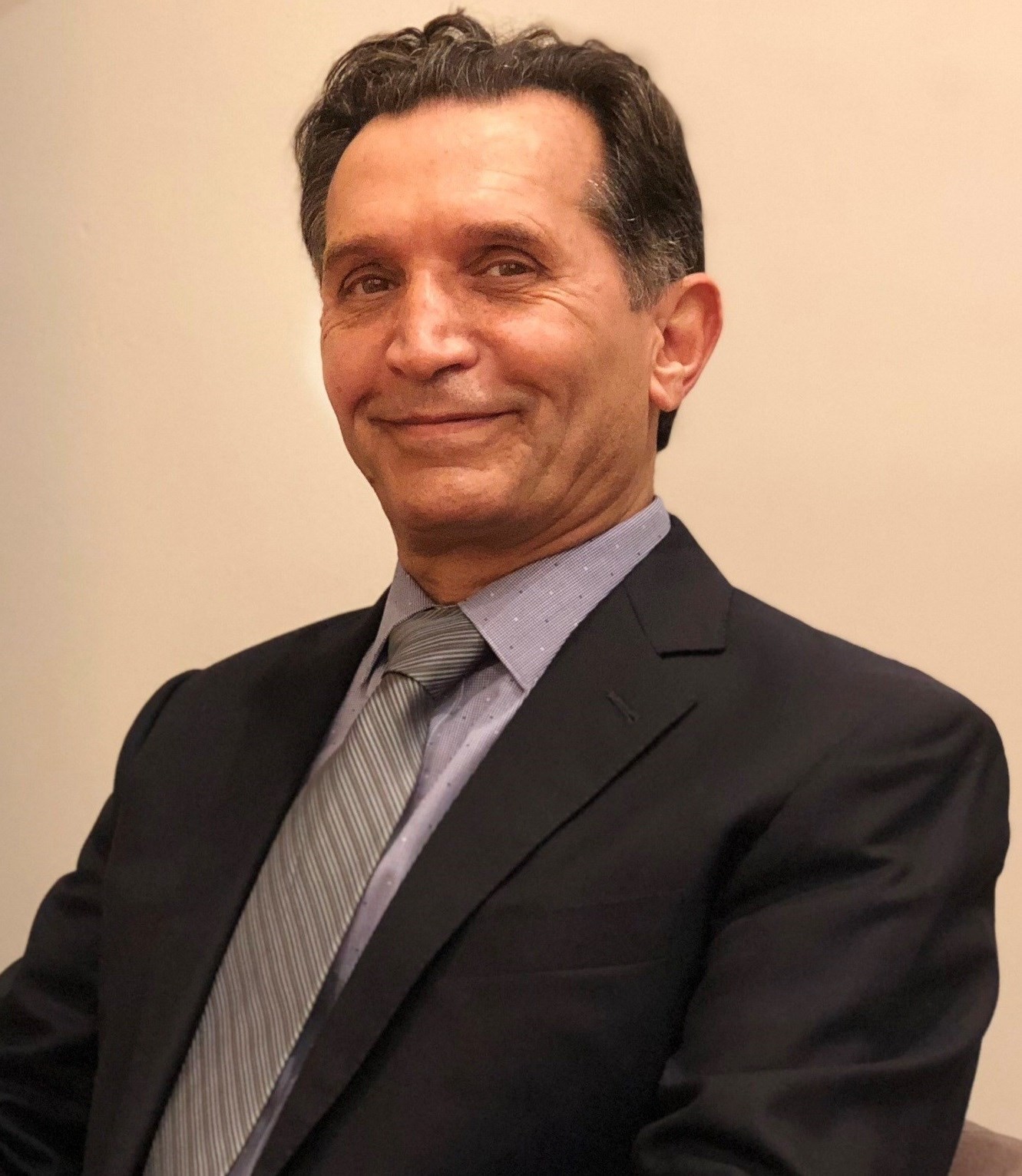}
\end{wrapfigure}
\textbf{Abdollah Homaifar} received his B.S. and M.S. degrees from the State University of New York at Stony Brook in 1979 and 1980, respectively, and his Ph.D. degree from the University of Alabama in 1987, all in Electrical Engineering. He is the NASA Langley Distinguished Professor and the Duke Energy Eminent professor in the Department of Electrical and Computer Engineering at North Carolina A$\&$T State University (NCA$\&$TSU). He is the director of the Autonomous Control and Information Technology Institute and the Testing, Evaluation, and Control of Heterogeneous Large-scale Systems of Autonomous Vehicles (TECHLAV) Center at NCA$\&$TSU. His research interests include machine learning, unmanned aerial vehicles (UAVs), testing and evaluation of autonomous vehicles, optimization, and signal processing. He also serves as an associate editor of the Journal of Intelligent Automation and Soft Computing and is a reviewer for IEEE Transactions on Fuzzy Systems, Man Machines and Cybernetics, and Neural Networks.  He is a member of the IEEE Control Society, Sigma Xi, Tau Beta Pi, and Eta Kapa Nu.

\begin{wrapfigure}{L}{0.2\columnwidth}
\centering
\includegraphics[width=0.2\columnwidth]{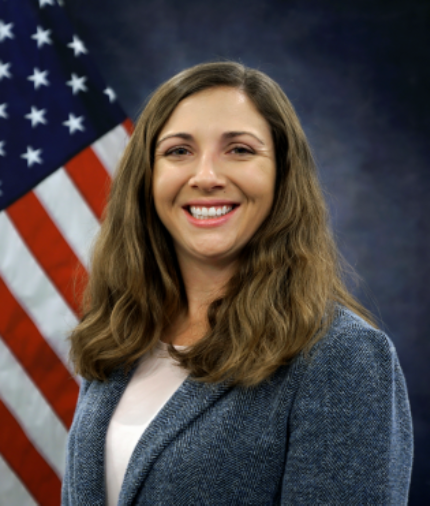}
\end{wrapfigure}
\textbf{Emily Doucette} serves as Multi-Domain Networked Weapons technical lead for the Air Force Research Laboratory Munitions Directorate. Prior to this post, she has served the Munitions Directorate as the Assistant to the
Chief Scientist (2017-2019) and as a research
engineer for the Weapon Dynamics and Control Sciences Branch since 2012. She earned
a Ph.D. in aerospace engineering from Auburn
University and is a recipient of the SMART
Scholarship. Her research interests include estimation theory, human-machine teaming, decentralized task assignment, cooperative autonomous engagement, and risk-aware target tracking and interdiction. Dr. Doucette leads a team of postdoctoral and graduate student researchers to support collaborative efforts across DoD,
industry, academia, and international partnerships. She served on the
AFRL Munitions Directorate Autonomy Steering Committee, is active in
the Autonomy Community of Interest, and is the co-lead for the OSD
Autonomy Center of Excellence. 

\end{document}